\documentclass[11pt]{article}

\usepackage[utf8]{inputenc}
\usepackage[T1]{fontenc}
\usepackage{amsmath,amssymb}
\usepackage{booktabs}
\usepackage{array}
\usepackage{caption}
\usepackage[letterpaper,margin=1in]{geometry}
\usepackage{xurl}
\usepackage[hidelinks]{hyperref}
\hypersetup{
  pdftitle={MemoryLake on MemoryArena: A Matched Study of Agent Memory Backends},
  pdfauthor={Chaoqun Zhan, Qiang Zhou, Guannan Li, Zhenqiang Huang, Qianjin Wang},
  pdfsubject={Matched system-level evaluation of agent memory backends on MemoryArena},
  pdfkeywords={agent memory, MemoryArena, long-term memory, LLM agents, benchmark evaluation}
}

\newcommand{\code}[1]{\texttt{#1}}

\title{MemoryLake on MemoryArena:\\ A Matched Study of Agent Memory Backends}
\author{Chaoqun Zhan \quad Qiang Zhou \quad Guannan Li \\ Zhenqiang Huang \quad Qianjin Wang \\[2pt]
  MemoryLake Team \\ \texttt{contact@zbyte-inc.com}}
\date{August 2026}

\begin{document}
\maketitle

\begin{abstract}
Most agent-memory benchmarks test post-hoc recall, whereas MemoryArena evaluates whether memory supports interdependent, multi-session task completion. We compare MemoryLake, a structured multi-track memory backend, with Mem0, \code{text-embedding-3-small} vector RAG, and a long-context control across all five MemoryArena domains. The systems share the same agent framework, requested \code{gpt-5-mini} model alias, task samples, and scoring code; the memory integration is the intentionally changed component. Because each backend bundles write, retrieval, consolidation, budgeting, and prompt-assembly choices, the study is a matched system-level comparison, not a representation-only ablation or a cost-matched experiment. On the shared evaluation sets, MemoryLake has the highest observed success rate (SR) in mathematics (9/40), physics (12/20), and progressive retrieval (4/20). Every system has zero SR in travel planning, and web shopping yields a single bundle-level success (long context, 1/150); MemoryLake ranks third on both the travel soft process score and shopping step match. Following MemoryArena's suite-level convention, a post-hoc equal-weight average over the five SRs is 20.5\% for MemoryLake versus 13.6\% for the best comparator. These are point estimates: sample sizes are modest, confidence intervals overlap, and we do not report paired significance tests. A separate MemoryLake-only run over all 221 progressive queries yields a failure-counted SR of 26.7\% (59/221) and is not a baseline comparison. The results support a workload-dependent view of memory backends and an observed lead among the four evaluated systems on the shared sets; they do not establish benchmark-wide state of the art or a causal advantage of representation structure.
\end{abstract}


\section{Introduction}
Agent-memory evaluation is moving from static recall toward task completion. Recall-oriented benchmarks such as LoCoMo, LongMemEval, MemoryAgentBench, and MemoryBench ask whether an agent can recover information from a long interaction history \cite{locomo,longmemeval,memoryagentbench,memorybench}. These benchmarks are useful, but they do not directly test whether remembered information changes later actions, survives error accumulation, or supports a final multi-step objective.

MemoryArena addresses this gap through a closed Memory--Agent--Environment loop \cite{memoryarena}. Each instance consists of ordered, interdependent subtasks. Information generated in an earlier subtask may become a premise, constraint, compatible product, or intermediate finding required later. The benchmark therefore separates process-level completion from end-to-end success: a system may recover some intermediate information without completing the final task.

We study MemoryLake in this setting. MemoryLake is a structured multi-track memory backend that maintains confirmed conclusions, supporting evidence, and reusable experience under different presence policies. We compare it with three controls supplied by or aligned with the public MemoryArena release: Mem0, naive vector-chunk RAG using \code{text-embedding-3-small}, and a same-model long-context configuration that flattens the prior trajectory into the prompt.

The study asks two questions. First, under a matched agent framework, model alias, task sample, and evaluator, which backend has the highest observed end-to-end success on the shared evaluation sets? Second, how does the ranking change across workloads that emphasize exact conclusion reuse, broad verbatim replay, constraint tracking, or aggregation of intermediate findings?

The main contributions are:
\begin{enumerate}
\item a matched system-level comparison of four memory configurations across all five MemoryArena domains, with exact sample sizes and denominators;
\item an end-to-end analysis that reports both per-domain SR and the benchmark's equal-weight suite-level average SR, while keeping process metrics visible;
\item a workload-level interpretation that distinguishes observed system performance from unsupported representation-level causality; and
\item a transparent account of sampling, judge dependence, artifact scope, and benchmark-release ambiguities that affect reproducibility.
\end{enumerate}

\paragraph{Claim scope.} The evidence supports a bounded claim: on the shared sets used in this study, MemoryLake has the highest observed SR on three domains and the highest post-hoc macro-average SR among the four evaluated systems. The evidence does not establish statistical significance, a benchmark-wide state-of-the-art result, resource-matched superiority, or a causal effect of any single representation mechanism.

\section{Related Work}
\paragraph{Recall-oriented memory benchmarks.} LoCoMo and LongMemEval evaluate question answering over long conversations \cite{locomo,longmemeval}. MemoryAgentBench and MemoryBench extend evaluation toward incremental interaction and continual learning \cite{memoryagentbench,memorybench}. Their common focus is whether information can be recovered after it has been stored; environment feedback and cross-subtask action dependencies are limited or absent.

\paragraph{Action-oriented agent benchmarks.} WebArena, WebShop, and SWE-bench evaluate navigation, tool use, grounded shopping, and software repair \cite{webarena,webshop,swebench}. These tasks test consequential actions, but instances are generally independent and do not require a persistent memory layer to transfer information across multiple sessions.

\paragraph{Memory--action coupled evaluation.} MemoryArena explicitly combines memory, action, and environment feedback across interdependent sessions \cite{memoryarena}. Its domains probe several memory demands: procedural reuse in formal reasoning, preference and constraint tracking in travel planning, exact item compatibility in bundled shopping, and aggregation of intermediate findings in progressive web search.

\paragraph{Memory systems.} Existing systems span extractive fact memory, trajectory retrieval, operating-system-style memory, graph-structured memory, and reusable reasoning or skill abstractions \cite{mem0,memgpt,graphrag,memorag,reasoningbank}. MemoryLake belongs to the structured-memory family but uses parallel tracks with different presence policies rather than a single homogeneous store.

Two recent works are particularly related. DeMem formulates memory as a decision-centric rate--distortion problem and reports MemoryArena results under a different model and protocol \cite{demem}. MSCE organizes grounded traces, procedural policies, and declarative cognition and promotes evidence-backed policies into callable skills \cite{msce}. Their published values are not treated as same-protocol baselines here: DeMem uses a different task backbone and evaluation configuration, while MSCE does not evaluate on MemoryArena.

\section{Evaluation Setup}
\subsection{Matched system-level protocol}
All four systems use the same MemoryArena agent framework, the same requested base-model alias (\code{gpt-5-mini}), the same task IDs for each cross-system comparison, and the benchmark's official scoring scripts, complemented by the progressive-retrieval aggregation described in Section~\ref{sec:scoring}. For MemoryLake, the same \code{gpt-5-mini} alias is used both for the task agent and for the backend's internal LLM operations; no larger or auxiliary language model is used inside the backend. Memory is reset between task instances. The benchmark environments and evaluators are not modified. Two changes to the pinned benchmark checkout are recorded: the agent's hard-coded output-token budget is raised identically for all four systems (Section~\ref{sec:threats}), and Mem0's progressive run applies a per-write size cap inside the Mem0 memory-service integration, gated on the backend name, without which that task cannot run to completion for Mem0 at all. The Mem0 change is system-specific; it is quantified in Section~\ref{sec:threats} and marked in every table where it appears. MemoryLake is connected through the same memory-service integration point used by the other external-memory configurations.

The integration point is the intentionally changed component, but a memory backend is not a single variable. It can change write policy, extraction, auxiliary-model calls, consolidation, retrieval, ordering, prompt assembly, context length, fallback behavior, and resource use. Accordingly, the experiment identifies the effect of assigning an end-to-end memory configuration under the tested setup. It is neither a representation-only ablation nor a token-, latency-, or cost-matched comparison.

\subsection{Systems}
The long-context system is a zero-abstraction control. Mem0 represents extractive fact memory. Vector RAG represents a simple embedding-based retrieval pipeline; among methods in the original MemoryArena table, \code{text-embedding-3-small} had the highest reported suite-level average SR \cite{memoryarena}. MemoryLake is described in Section~\ref{sec:memorylake}, including its use of the \code{gpt-5-mini} language model and the \code{bge-m3} dense embedder.

\begin{table}[t]
\centering
\small
\caption{System-level differences. The experiment compares complete backend configurations rather than isolating representation structure.}
\label{tab:systems}
\begin{tabular}{@{}p{2.4cm}p{5.0cm}p{5.0cm}@{}}
\toprule
System & Stored representation & Recall and prompt behavior \\
\midrule
Long Context & No external memory; prior within-instance trajectory retained verbatim & No retrieval; full available trajectory is flattened into the next prompt \\
Mem0 & Extractive fact entries & Similar-fact recall through the Mem0 backend \\
Vector RAG & Raw trajectory chunks embedded with \code{text-embedding-3-small} & Similar-chunk retrieval \\
MemoryLake & Confirmed conclusions, supporting evidence, and reusable experience in separate tracks & Prioritized conclusion presence, on-demand evidence retrieval (dense encoding via \code{bge-m3}), and bounded assembly \\
\bottomrule
\end{tabular}
\end{table}

\subsection{Tasks, samples, and metrics}
Table~\ref{tab:tasks} summarizes the common evaluation sets. The cross-system claims in this paper use only IDs shared by all four systems.

\paragraph{Formal mathematical reasoning.} The released set contains 40 paper-level instances decomposed into 354 ordered subproblems. Each paper is one memory session; the conclusion of an earlier step may be required later. The official evaluator uses an LLM-based mathematical-equivalence judgment. PS is the mean, across papers, of within-paper subproblem accuracy. SR is the fraction of papers whose final subproblem is judged correct. We use the full released set.

\paragraph{Formal physical reasoning.} The physical-reasoning set contains 20 papers and 86 ordered subproblems. The protocol and metrics mirror the mathematical task. We use the full released set.

\paragraph{Group travel planning.} The released benchmark contains 270 groups. We use a fixed list of 30 groups, comprising 208 member-level plans, identical across systems. The evaluator performs deterministic slot matching. SPS measures satisfaction of the added member constraints relative to the base itinerary. SR requires every member in a group to pass.

\begin{table}[t]
\centering
\small
\caption{Shared comparison sets and metrics. Travel and progressive retrieval use subsets of the released benchmark.}
\label{tab:tasks}
\begin{tabular}{@{}p{3.0cm}p{2.8cm}p{3.4cm}p{3.4cm}@{}}
\toprule
Domain & Shared scale & Process metric & End-to-end metric \\
\midrule
Formal mathematics & 40 papers / 354 subproblems & Mean within-paper subproblem accuracy (PS) & Final subproblem correct for each paper (SR) \\
Formal physics & 20 papers / 86 subproblems & Mean within-paper subproblem accuracy (PS) & Final subproblem correct for each paper (SR) \\
Group travel planning & 30 groups / 208 member plans & Added-constraint slot satisfaction (SPS) & Every member plan in a group passes (SR) \\
Bundled web shopping & 150 bundles / 900 purchases & Exact ASIN step match & All six purchases in a bundle match (SR) \\
Progressive retrieval & 20 queries / 142 slots & All-slots macro-average pass rate (PS) & Final combined query correct (SR) \\
\bottomrule
\end{tabular}
\end{table}

\paragraph{Bundled web shopping.} Each bundle contains six sequential purchases with cross-step compatibility and budget constraints. All four systems completed the full released set of 150 bundles (5 categories $\times$ 30 bundles, 900 steps), and the cross-system table scores that complete set. A fixed 50-bundle, 300-step sensitivity subset --- the first 10 bundles of each category --- is additionally scored to check the stability of the step-match ordering across scales.

\paragraph{Progressive retrieval.} The released data contain 221 queries with progressive decompositions, totaling 1{,}641 slots (1{,}420 sub-queries and 221 final combined queries). The main comparison uses a fixed subset of 20 queries and 142 slots (122 sub-queries and 20 final queries), selected before scoring and shared by all systems. The subset is not a random sample: it is a deterministic proportional stratified sample on decomposition depth, drawn with no random seed. The 221 decomposed queries are stratified by slot count, each stratum receives a quota of $\mathrm{round}(n_{\text{stratum}} \times 20/221)$, and the lowest query IDs within a stratum fill that quota. Depth is the variable this task is built to stress, since a conclusion must survive every step of the chain, so sampling proportionally on it keeps the subset's difficulty profile matched to the full 221 rather than skewed toward short or long chains. The original selection script was not preserved; the rule stated here was recovered from the published subset and regenerates it exactly, so it is a verified reconstruction rather than the original code, and the subset can be rebuilt offline from the published manifest. The same environment-failure handling is applied across systems. A separate MemoryLake-only run covers all 221 released progressive queries; it is reported separately and is not used for cross-system ranking.

\paragraph{Progressive PS and SR definitions.} For each query $q$, let $S_q$ denote its slot count (sub-queries plus the final combined query), $a_q \le S_q$ the slots that produced a non-empty answer, and $p_q$ the number judged correct. We report $\mathrm{PS} = \frac{1}{Q}\sum_q p_q / S_q$, the macro-average over queries of the all-slots pass rate; unanswered slots count as incorrect. This differs from the scorer's default, which divides by $a_q$ and therefore measures a system that answers fewer slots on an easier denominator. Slot coverage $\sum_q a_q / \sum_q S_q$ is reported alongside PS in Table~\ref{tab:coverage}. Both forms are macro-averages over queries rather than $\sum_q p_q / \sum_q S_q$ over slots, so a 2-slot query carries the same weight as a 16-slot one. Because $S_q$ includes the final combined query, that item enters PS as well as SR and the two measures are not independent.

\subsection{Scoring, suite-level summary, and verification}
\label{sec:scoring}
MemoryArena's official scoring scripts produce the mathematics, physics, travel, and shopping metrics. The benchmark ships no end-to-end scorer for progressive retrieval: it provides the LLM judge but not the per-query PS/SR aggregation over slots. That task therefore uses the benchmark's judge together with an aggregation script we supply, applied identically to all four systems and released with the artifact. Travel and shopping are deterministic. Mathematics, physics, and progressive retrieval rely on the benchmark's LLM-based judging configuration.

Following the suite-level summary used by MemoryArena, we also report the equal-weight average of the five domain SR values:
\begin{equation}
\mathrm{MacroAvgSR} = \frac{1}{5}\sum_{d=1}^{5} \mathrm{SR}_d .
\end{equation}
This average is useful because SR is the common end-to-end outcome across domains. It is nevertheless post hoc in this study, was not preregistered, and gives equal weight to domains with different sample sizes and partial-coverage subsets. We therefore treat it as a descriptive summary rather than an inferential primary endpoint.

A second implementation was used internally to recompute aggregate metrics from the authors' stored result files. Deterministic tasks matched exactly. The authors report sub-one-percentage-point aggregate differences on judged tasks. Because this check is not a fresh independent judging pass, we treat it as an arithmetic consistency check rather than independent replication; the per-instance scored outputs underlying it are included in the public artifact so that the aggregate arithmetic can be re-verified externally.

\subsection{Statistical reporting}
SR is a binary per-instance rate, so Table~\ref{tab:counts} reports exact counts and Table~\ref{tab:wilson} reports marginal 95\% Wilson intervals for all five comparison domains. These intervals describe each system separately; they are not substitutes for a paired comparison. The present version does not report exact McNemar tests for SR or clustered paired bootstrap intervals for PS, SPS, and step match. All rank statements are therefore explicitly point-estimate statements, not claims of statistical significance.

\section{MemoryLake: Structured Multi-Track Memory}
\label{sec:memorylake}
MemoryLake connects through two generic operations. After each completed subtask, the agent writes the trajectory and outcome to memory. Before the next subtask, MemoryLake returns an assembled memory context that is prepended to the new question. The same generic configuration is used across all five domains; no task-specific tuning is reported.

\paragraph{Implementation components.} All LLM operations in MemoryLake --- the task agent together with the internal consolidation, extraction, and assembly steps --- use the same requested \code{gpt-5-mini} alias; no larger or auxiliary language model is used inside the backend. Dense retrieval uses the open BAAI \code{bge-m3} encoder in dense-only mode. Both the language model and the embedder are therefore publicly available, standard components; what remains proprietary is the multi-track storage, indexing, consolidation, and assembly engine described below.

\subsection{Presence policies for heterogeneous content}
MemoryLake separates three kinds of content. First, confirmed conclusions from completed subtasks are maintained in order and prioritized for presence in later subtasks. This reduces reliance on semantic similarity for information that is already known to be a prerequisite. Second, supporting evidence and procedural traces are stored for on-demand retrieval. Third, reusable problem-solving experience is consolidated for transfer across subtasks.

At recall time, the tracks are assembled under a bounded context budget. Redundant material may be filtered or compacted, and not all historical content is retained verbatim. If the internal memory service fails, the integration returns empty memory context and allows the agent to continue rather than blocking the task.

\subsection{Disclosure scope}
The paper describes MemoryLake at the representation and policy level and identifies its externally visible components: the \code{gpt-5-mini} language model and the \code{bge-m3} dense embedder, both publicly available. Its storage layout, indexing, consolidation, and assembly heuristics remain proprietary. The evaluation artifacts support inspection of task IDs, protocols, and released scored outputs, but they cannot provide full independent reproduction of the MemoryLake backend. We therefore distinguish score verification from backend reproducibility.

\subsection{Design hypothesis}
The design hypothesis is that different memory content benefits from different presence policies: confirmed conclusions may require prioritized availability, supporting evidence may be best retrieved on demand, and reusable procedures may benefit from consolidation. The experiment does not isolate any of these mechanisms. Results are interpreted as evidence about the complete backend configuration, and mechanism-level explanations remain hypotheses requiring ablation.

\section{Results}
\subsection{Domain-level results}
MemoryLake has the highest observed SR in mathematics, physics, and progressive retrieval. It leads neither the travel nor the shopping process metric. Travel yields zero SR for every system, and shopping yields a single bundle-level success across all four systems (long context, 1/150); neither domain distinguishes end-to-end success under the present sample and strict success definition.

\subsection{End-to-end SR summary}
On this descriptive suite-level summary, MemoryLake reaches 20.5\% macro-average SR, 6.9 percentage points above the best comparator value of 13.6\%. The difference is driven by the formal-reasoning and progressive-retrieval domains; travel contributes zero to every system, and shopping contributes at most one success (long context, 1/150). Because the endpoint was not preregistered and two domains use subsets, it should be read as a compact summary of this experiment, not as a benchmark-wide leaderboard result.

\subsection{Per-domain interpretation}
\paragraph{Physics.} Physics is MemoryLake's largest observed SR margin: 12/20 versus 9/20 for long context. The task has relatively short chains but requires exact reuse of intermediate derivations. The result is consistent with prioritized conclusion presence being useful, but the experiment does not isolate that mechanism and the marginal intervals overlap.

\paragraph{Mathematics.} Mathematical PS is similar across systems, ranging from 27.6\% to 29.5\%, while final SR ranges from 12.5\% to 22.5\%. MemoryLake's advantage over Mem0 is one paper (9/40 versus 8/40). The appropriate interpretation is a highest observed point estimate, not an established margin.

\paragraph{Progressive retrieval.} This domain shows the clearest process--outcome rank reversal. Mem0 has the highest all-slots PS (8.9\%), but MemoryLake has the highest final-query SR (4/20 versus 3/20 for Mem0 and 2/20 for each other system). Mem0's figures are those of a capped re-run rather than its unmodified run, which never reached the final combined query and scored 0/20; the reason and both sets of values are given in Section~\ref{sec:threats}. Vector RAG shows why the PS denominator matters: it reads 8.2\% under the scorer's default answered-slot denominator but 5.6\% on the all-slots denominator used here, because it answers only 64.8\% of its slots against MemoryLake's 93.7\%. A hypothesis consistent with MemoryLake's SR lead is that its grouped assembly helps expose intermediate findings at final synthesis time. Because no assembly ablation is included, the result cannot be attributed to that mechanism alone.

\paragraph{Web shopping.} On the complete 150-bundle set, long context has the highest step match at 30.0\%, followed by vector RAG at 29.7\%, MemoryLake at 29.6\%, and Mem0 at 24.3\%. The single bundle-level success in this domain belongs to long context (1/150); the other three systems score 0/150. The three leading systems fall within 0.4 percentage points over 900 steps, which this sample cannot resolve as a separation, so the step-match column should be read as no system separating rather than as a ranking. The 50-bundle sensitivity subset reorders the leaders --- vector RAG 31.0\%, MemoryLake 30.0\%, long context 28.3\%, Mem0 24.0\%, with zero bundle-level SR for every system --- confirming that the step-match ordering is not stable at these sample sizes. MemoryLake does not lead this workload at either scale.

\paragraph{Travel planning.} Long context leads SPS at 53.1\%, followed by vector RAG at 45.7\%, MemoryLake at 43.0\%, and Mem0 at 27.2\%. This ordering is consistent with a fidelity--abstraction trade-off for high-volume slot-by-slot replay. It also marks a clear boundary of the current MemoryLake configuration: structured assembly does not provide the best process score on this workload, and every system has zero group-level SR.

\begin{table}[t]
\centering
\small
\caption{Main results in percent. Bold indicates the highest point estimate for each displayed metric. SR is zero for every system in travel planning; in web shopping the only bundle-level success is long context's 1/150.}
\label{tab:main}
\begin{tabular}{@{}llcccc@{}}
\toprule
Task & Metric & MemoryLake & Mem0 & Vector RAG & Long Context \\
\midrule
Math reasoning & PS / SR & \textbf{29.5} / \textbf{22.5} & 28.4 / 20.0 & 28.9 / 17.5 & 27.6 / 12.5 \\
Physics reasoning & PS / SR & \textbf{64.5} / \textbf{60.0} & 45.9 / 30.0 & 57.4 / 40.0 & 59.1 / 45.0 \\
Travel planning & SPS / SR & 43.0 / 0.0 & 27.2 / 0.0 & 45.7 / 0.0 & \textbf{53.1} / 0.0 \\
Web shopping & Step match / SR & 29.6 / 0.0 & 24.3 / 0.0 & 29.7 / 0.0 & \textbf{30.0} / \textbf{0.7} \\
Progressive retrieval & PS / SR & 6.7 / \textbf{20.0} & 8.9 / 15.0$^{*}$ & 5.6 / 10.0 & 5.3 / 10.0 \\
\bottomrule
\end{tabular}

\vspace{4pt}
\parbox{0.97\textwidth}{\footnotesize $^{*}$Mem0's progressive figures are those of a re-run in which each memory write is truncated to a per-write size cap inside the Mem0 integration; no other system received this change, so this is the one cell not measured under identical code. Its 8.9 is the highest progressive PS but is left unbolded for that reason: it is not a like-for-like measurement. Mem0's unmodified run reached the final combined query on 0 of 20 queries (SR 0/20, all-slots PS 1.9\%). Both runs are released with the artifact; the reason for reporting the capped run is given in Section~\ref{sec:threats}.}
\end{table}

\begin{table}[t]
\centering
\small
\caption{Exact end-to-end success counts and the post-hoc equal-weight average of the five domain SRs. The average is descriptive and is computed on the shared sets, not the complete benchmark. Mem0's starred progressive count is the capped re-run defined under Table~\ref{tab:main}; its unmodified run scores 0/20, which lowers its average to 10.0\%.}
\label{tab:counts}
\begin{tabular}{@{}lcccccc@{}}
\toprule
System & Math & Physics & Travel & Shopping & Progressive & Macro Avg SR \\
\midrule
MemoryLake & 9/40 & 12/20 & 0/30 & 0/150 & 4/20 & \textbf{20.5\%} \\
Mem0 & 8/40 & 6/20 & 0/30 & 0/150 & 3/20$^{*}$ & 13.0\% \\
Vector RAG & 7/40 & 8/20 & 0/30 & 0/150 & 2/20 & 13.5\% \\
Long Context & 5/40 & 9/20 & 0/30 & 1/150 & 2/20 & 13.6\% \\
\bottomrule
\end{tabular}
\end{table}

\subsection{MemoryLake-only full progressive run}
The shared 20-query progressive sample is small, so MemoryLake was additionally run on all 221 progressive decompositions in the released dataset. Its failure-counted SR is 26.7\% (59/221; 95\% Wilson interval [21.3, 32.9]) and its all-slots PS is 12.3\% (13.4\% on the scorer's answered-slot denominator). Instances that did not complete under the run configuration, including input-length failures, are counted as failures. The reported values are those of a slot-level merge of the original run with re-runs of the slots it left unanswered: each slot takes its answer from the first source that produced one, so a re-run can only fill an empty slot and can never replace an answer that was already scored. The merge raises slot coverage from 1315/1641 (80.1\%) to 1538/1641 (93.7\%), the same coverage as the shared 20-query subset, and SR from 16.7\% (37/221) to 26.7\% (59/221); both values are stated because the headline figure is a post-merge one. The 103 slots (6.3\%) still unanswered are runs that reached the search-iteration cap without emitting an answer. First-wins rather than last-wins is deliberate: scoring the same 20 queries twice under different output budgets produced SR 20.0\% and 15.0\%, so preferring the newer answer could have replaced correct answers with incorrect ones. This result estimates MemoryLake's own performance on the full released progressive set. It is not compared with baseline values from the 20-query subset and does not establish a cross-system advantage at the 221-query scale.

\begin{table}[t]
\centering
\small
\caption{Exact SR fractions and marginal Wilson intervals for all five comparison domains. Rows with identical results across systems are reported in compact form. Overlapping marginal intervals do not constitute a paired significance test. The 221-query row is MemoryLake-only and is not comparable to the 20-query baselines. Mem0's starred progressive row is the capped re-run defined under Table~\ref{tab:main}; its unmodified run is 0/20 (0.0\%), 95\% Wilson interval [0.0, 16.1].}
\label{tab:wilson}
\begin{tabular}{@{}llcc@{}}
\toprule
Task & System & SR & 95\% Wilson interval \\
\midrule
Math ($n=40$) & MemoryLake & 9/40 (22.5\%) & [12.3, 37.5] \\
 & Mem0 & 8/40 (20.0\%) & [10.5, 34.8] \\
 & Vector RAG & 7/40 (17.5\%) & [8.7, 32.0] \\
 & Long Context & 5/40 (12.5\%) & [5.5, 26.1] \\
\midrule
Physics ($n=20$) & MemoryLake & 12/20 (60.0\%) & [38.7, 78.1] \\
 & Mem0 & 6/20 (30.0\%) & [14.5, 51.9] \\
 & Vector RAG & 8/20 (40.0\%) & [21.9, 61.3] \\
 & Long Context & 9/20 (45.0\%) & [25.8, 65.8] \\
\midrule
Travel ($n=30$) & All four systems & 0/30 (0.0\%) & [0.0, 11.4] \\
Shopping ($n=150$) & All but Long Context & 0/150 (0.0\%) & [0.0, 2.5] \\
 & Long Context & 1/150 (0.7\%) & [0.1, 3.7] \\
\midrule
Progressive ($n=20$) & MemoryLake & 4/20 (20.0\%) & [8.1, 41.6] \\
 & Mem0$^{*}$ & 3/20 (15.0\%) & [5.2, 36.0] \\
 & Vector RAG & 2/20 (10.0\%) & [2.8, 30.1] \\
 & Long Context & 2/20 (10.0\%) & [2.8, 30.1] \\
\midrule
Progressive, ML only ($n=221$) & MemoryLake & 59/221 (26.7\%) & [21.3, 32.9] \\
\bottomrule
\end{tabular}
\end{table}

\section{Discussion and Limitations}
\subsection{What the experiment establishes}
The observed ranking is workload dependent. MemoryLake leads the end-to-end point estimate in formal reasoning and progressive retrieval, while long context leads both the travel and shopping process metrics. This pattern argues against treating ``memory system'' as a single universally ordered category. It also shows why process metrics and final success should be reported together: a backend can retrieve or solve more intermediate items without having the highest final success, and vice versa.

The suite-level macro-average SR gives a direct summary aligned with the study's end-to-end motivation. On the shared sets, MemoryLake's 20.5\% is higher than the 13.6\% best-comparator value. This is the strongest single descriptive statement supported by the current tables. It remains bounded by the subset design, the absence of paired significance tests, and incomplete same-protocol baseline coverage.

\subsection{Threats to validity}
\label{sec:threats}
\paragraph{Small samples and no paired inference.} The math margin over Mem0 is one instance, the progressive margin over Mem0 is one instance, and the physics margin over long context is three instances. Marginal Wilson intervals overlap. Exact paired tests and clustered paired bootstrap intervals are not reported, so no statistical-win claim is made.

\paragraph{Partial and non-random coverage.} Travel uses 30 of 270 groups and the progressive comparison uses 20 of 221 queries; mathematics, physics, and shopping use the complete released sets. The travel and progressive subsets may not represent the complete benchmark, and the post-hoc macro-average SR therefore mixes three full-domain runs with two subset runs. The 50-bundle shopping sensitivity check also reorders the step-match leaders relative to the complete set, which underlines how little weight the sub-percentage-point step-match differences can bear.

\paragraph{Treatment versus mechanism.} The backend swap changes several mechanisms at once. The results cannot identify whether any difference is caused by representation, retrieval, consolidation, prompt assembly, fallback behavior, or another backend property. Mechanism claims require targeted ablation.

\paragraph{Embedding-quality confound.} MemoryLake performs dense retrieval with \code{bge-m3}, whereas the Vector RAG baseline uses \code{text-embedding-3-small}; the two backends therefore differ in embedder as well as in memory structure. We did not run an embedder-matched control and do not quantify the embedder's separate contribution. We note, however, that this confound is unlikely to favor MemoryLake. The distinctive strengths of \code{bge-m3} --- multilingual coverage, hybrid dense--sparse--multi-vector scoring, and long-document granularity --- are inactive in MemoryArena's monolingual English, short-chunk, dense-only regime, and the original MemoryArena evaluation reports \code{text-embedding-3-small} as its strongest embedding baseline \cite{memoryarena}. Under these conditions MemoryLake's dense-only \code{bge-m3} is not expected to hold an embedding-quality advantage over the baseline's \code{text-embedding-3-small}, so any residual embedding effect would, if anything, work against MemoryLake rather than explain its observed lead. This is a directional argument from the embedders' design profiles and the benchmark's published embedder ranking, not a same-benchmark head-to-head; a quantitative attribution would require an embedder-matched run, which we leave to future work.

\paragraph{Resource fairness.} The study does not match or comprehensively report token consumption, auxiliary-model calls, latency, or cost. It should be interpreted as a quality comparison under each backend's tested configuration, not as an efficiency comparison or a conclusion about equal-resource performance.

\paragraph{Final-query search-iteration cap.} In the progressive task the sub-query agent receives the configured search-iteration cap, 35 in our runs, but the final-query entry point omits the argument, so the final combined query silently falls back to the agent's built-in default of 30. Because SR is computed solely from the final query, the most frequently quoted metric is the one measured under the smaller budget. Across our 386 final-query runs, 90 (23\%) produced no answer and 79 of those stopped at exactly 30 tool calls. The effect is unlikely to be large --- among runs that did answer, the median used 7 tool calls and only 2 of 297 answered at the cap --- and we estimate a 1--4 percentage-point SR effect at $n=221$, but the confound is real. It applies identically to all four systems, so it does not favor any of them.

\paragraph{Non-configurable agent output budget.} The agent's output-token budget is hard-coded rather than exposed in any configuration file; our runs use 32{,}000 against an upstream value of 15{,}000. A run reconstructed from the public configuration alone will therefore not reproduce these scores, which is a further reason a tagged run manifest is needed.

\paragraph{Dropped samples rather than low scores.} When the progressive agent returns no answer, the upstream code serializes the entire run object, approximately 9\,MB, into the answer field and sends it to the judge; the judge exceeds its context window and returns 400, the environment step returns 500, and the query disappears from the results rather than scoring zero. A failure of this kind removes a sample instead of lowering a score and is invisible in a score table, which is why slot coverage is reported alongside PS.

\paragraph{Model-version reproducibility.} The requested model alias is \code{gpt-5-mini}. Resolved snapshot IDs, complete run dates, and all decoding and retry settings are not available in the current public artifact. Hosted-model behavior may therefore be difficult to reproduce exactly at a later date. A tagged run manifest is needed for stronger temporal reproducibility.

\paragraph{Judge dependence.} Three domains rely on LLM-based judgments. The internal recomputation checks aggregate arithmetic but does not independently validate judge accuracy or stability. The public release includes the stored per-instance scored outputs, which allow the reported labels to be inspected; inter-judge agreement statistics and a blind re-audit of ranking-critical instances remain future additions.

\paragraph{Mem0's per-write size limit and the capped re-run.} Mem0's progressive figures are those of a re-run in which each memory write is truncated to a per-write size cap, and this is the study's one system-specific code change. The unmodified run does not measure recall at all: the environment writes the full agent trace into memory after each sub-query --- median 58{,}217 tokens over 132 writes, maximum 216{,}751 --- while the Mem0 cloud API rejects any single write above 100{,}000 tokens, so the write raises, the environment step returns 500, and the query is skipped after three retries. Mem0 answered 32 of 142 slots, reached the final combined query on 0 of 20 queries, and therefore scored SR 0/20 with an all-slots PS of 1.9\%; because SR is computed solely from the final query, no other value was reachable. The cap moves Mem0 from 0/20 to 20/20 on reaching the final query, its SR to 15.0\% (3/20) and its all-slots PS to 8.9\%, which is what Tables~\ref{tab:main}--\ref{tab:wilson} report. We prefer the capped run because 0/20 measures an API size limit rather than extractive fact memory, but the choice cuts both ways: the cap is gated on the backend name and no other system received it, so Mem0's progressive row is not measured under the same code as the others, and its 8.9\% PS lead in Table~\ref{tab:main} in particular should not be read as a like-for-like result. Both runs are released with the artifact, and Table~\ref{tab:counts} states the macro-average under either choice. Two further caveats attach to the capped run: it retains roughly the first third of each trace (30.3\% of total content across the run), and Mem0 scores better on that third than on the whole trace, which suggests writing a verbatim search transcript into memory may be actively harmful rather than merely wasteful --- a hypothesis, since the same cap was not applied to the other three systems.

\paragraph{Baseline coverage.} The comparison includes one extractive memory system, one simple vector-RAG system, and long context. It does not include same-protocol runs of other structured memory systems such as Letta, GraphRAG, ReasoningBank, or DeMem. DeMem reports stronger MemoryArena results under a different backbone and setup \cite{demem}; those values cannot be merged with the present table, but they preclude a benchmark-wide state-of-the-art claim from this experiment alone.

\paragraph{Proprietary implementation.} MemoryLake's internal implementation is not released. Public artifacts can verify protocols and scores, but cannot reproduce the backend itself. The commercial affiliation of all authors also creates an incentive for favorable interpretation, which is addressed through explicit point-estimate language and disclosure rather than eliminated.

\paragraph{Generalization.} Results are limited to the tested model tier, benchmark release, and task configurations. Stronger or weaker base models, alternative prompts, different context budgets, or other agent frameworks may change the relative ranking.

\subsection{Alignment with the MemoryArena v1 release}
The MemoryArena paper and public release contain several ambiguities relevant to reproduction. We record only those that affect this study's protocol: the model identifier is not used consistently in the paper; the paper cites 256 progressive tasks while the released data contain 221 progressive decompositions; and the original progressive numbers could not be reconstructed under our interpretation of the public sampling and evaluator. Appendix~\ref{app:alignment} describes how we handle these points. None is used to adjust the within-experiment scores in Tables~\ref{tab:main}--\ref{tab:wilson}.

\section{Conclusion}
We compared MemoryLake, Mem0, vector RAG, and long context across five MemoryArena domains under a matched agent framework, requested model alias, shared task IDs, and shared scoring code. On the shared sets, MemoryLake has the highest observed SR in mathematics (9/40), physics (12/20), and progressive retrieval (4/20). Mem0 reaches 3/20 on progressive retrieval, while vector RAG and long context each reach 2/20; Mem0's progressive run required a per-write size cap inside its integration, without which it never reached the final combined query and scored 0/20. In travel every system has zero end-to-end SR, and in shopping the only success is long context's 1/150; MemoryLake ranks third on both travel SPS and shopping step match.

The post-hoc equal-weight average across the five SRs is 20.5\% for MemoryLake and 13.6\% for the best comparator. This is a concise statement of the observed end-to-end result among the four evaluated systems on the shared sets. It is not evidence of benchmark-wide state of the art: the comparison uses partial subsets in two domains, does not include all relevant structured-memory baselines, does not match resource budgets, and does not report paired significance tests.

Taken together, the results support a workload-dependent view of memory backends. MemoryLake's current configuration is most promising where later decisions depend on exact conclusions or aggregation of intermediate findings, whereas long context remains competitive when the task rewards broad, high-fidelity replay. Establishing a stronger leadership claim will require full common-set runs, a pinned model snapshot, paired inference, resource logging, and a complete public result artifact.

\section*{Competing Interests}
All authors are affiliated with MemoryLake, a commercial memory system, and the compared systems are third-party baselines. The study uses the benchmark's official evaluators and reports negative as well as positive results, but these safeguards do not remove the conflict of interest. Readers should interpret mechanism explanations as hypotheses and system rankings as bounded by the stated protocol.

\section*{Code and Data Availability}
The MemoryArena environments and evaluators are available from the benchmark authors \cite{memoryarena}. The companion repository for this study is \url{https://github.com/memorylake-ai/memorylake-memoryarena-benchmark}. It contains the method description, evaluation settings, exact sample manifests, the per-instance scored outputs, and the aggregation scripts sufficient to recompute the aggregate values in Tables~\ref{tab:main}--\ref{tab:wilson} from the stored per-instance results. The released artifacts support verification of task selection, scoring inputs, and aggregate arithmetic. They do not re-execute the benchmark's LLM-based judges: judged-task labels are verified as released rather than regenerated, so score verification is distinct from independent re-judging. MemoryLake's internal storage, indexing, consolidation, and assembly engine remains proprietary and is not released.

\appendix
\section{Alignment with the MemoryArena v1 Release}
\label{app:alignment}

\begin{table}[h]
\centering
\small
\caption{Release-alignment observations that directly affect protocol interpretation.}
\label{tab:alignment}
\begin{tabular}{@{}p{3.0cm}p{5.0cm}p{5.0cm}@{}}
\toprule
Issue & Observed release ambiguity & Treatment in this study \\
\midrule
Model identifier & The MemoryArena paper uses both GPT-5.1-mini and GPT-5-mini in different sections. & We report the requested alias used in our runs, \code{gpt-5-mini}, and do not use cross-paper absolute score matching. \\
Progressive count & The paper cites 256 progressive tasks, while the released dataset contains 221 progressive decompositions. & The shared comparison uses a fixed 20-query list. The separate full MemoryLake run uses all 221 released decompositions. \\
Progressive reproduction & The paper does not provide a reproducible query list and evaluator version sufficient for us to reconstruct its reported 20-query values. & We use only scores generated inside our four-system experiment and do not import the original paper's progressive values. \\
\bottomrule
\end{tabular}
\end{table}

These notes are not presented as a re-evaluation of the MemoryArena contribution. They explain why all quantitative claims in this paper are within-experiment claims based on shared samples and evaluators.

\section{Scoring and Verification Notes}
\label{app:scoring}
Math and physics use the benchmark's LLM-based equivalence judging. Travel uses deterministic slot matching. Shopping uses exact ASIN matching. Progressive retrieval uses the benchmark's LLM-based final-answer judge together with the slot-level PS/SR aggregation described in Section~\ref{sec:scoring}, which the benchmark does not ship. The exact cross-system SR denominators are 40 papers for mathematics, 20 papers for physics, 30 groups for travel, 150 bundles for shopping, and 20 queries for progressive retrieval.

Table~\ref{tab:coverage} reports progressive slot coverage together with both PS denominators, and both Mem0 runs. Coverage varies more than fourfold across systems, which is why the cross-system PS in Table~\ref{tab:main} uses the all-slots denominator: the scorer's default skips unanswered slots instead of scoring them incorrect, so it measures a system that answers fewer slots on an easier denominator.

\begin{table}[h]
\centering
\small
\caption{Progressive retrieval on the shared 20-query subset: slot coverage and both PS denominators, in percent unless a fraction is shown. Cross-system comparisons use the all-slots denominator. Both Mem0 runs are listed; no cell is bolded, because the capped re-run is not measured under the same code as the other rows.}
\label{tab:coverage}
\begin{tabular}{@{}lcccc@{}}
\toprule
System & Slot coverage & PS (answered slots) & PS (all slots) & SR \\
\midrule
MemoryLake & 133/142 (93.7) & 6.7 & 6.7 & 4/20 \\
Mem0, unmodified & 32/142 (22.5) & 3.3 & 1.9 & 0/20 \\
Mem0, capped re-run & 120/142 (84.5) & 10.7 & 8.9 & 3/20 \\
Vector RAG & 92/142 (64.8) & 8.2 & 5.6 & 2/20 \\
Long Context & 134/142 (94.4) & 6.0 & 5.3 & 2/20 \\
\bottomrule
\end{tabular}
\end{table}

The independent aggregation program is an internal consistency check over the authors' stored results. Exact agreement on deterministic tasks provides evidence that the denominators and formulas were implemented consistently. The reported sub-one-percentage-point differences on judged tasks are not characterized as independent replication because the current release does not document whether all differences arise from parsing, rounding, or judge-output handling. The released artifact preserves, for each scored instance, the parsed label and the final aggregate row, together with the aggregation scripts, so that Tables~\ref{tab:main}--\ref{tab:wilson} can be recomputed from the stored results. Raw agent transcripts and full evaluator input--output are included where available; complete verbatim judge transcripts and inter-judge agreement statistics remain future additions.


\begin{thebibliography}{99}

\bibitem{locomo} A. Maharana, D.-H. Lee, S. Tulyakov, M. Bansal, F. Barbieri, and Y. Fang. Evaluating very long-term conversational memory of LLM agents. In \emph{Proceedings of the 62nd Annual Meeting of the Association for Computational Linguistics}, 2024.

\bibitem{longmemeval} D. Wu, H. Wang, W. Yu, Y. Zhang, K.-W. Chang, and D. Yu. LongMemEval: Benchmarking chat assistants on long-term interactive memory. In \emph{International Conference on Learning Representations}, 2025.

\bibitem{memoryagentbench} Y. Hu, Y. Wang, and J. McAuley. Evaluating memory in LLM agents via incremental multi-turn interactions. \emph{arXiv preprint arXiv:2507.05257}, 2025.

\bibitem{memorybench} Q. Ai, Y. Tang, C. Wang, J. Long, W. Su, and Y. Liu. MemoryBench: A benchmark for memory and continual learning in LLM systems. \emph{arXiv preprint arXiv:2510.17281}, 2025.

\bibitem{memoryarena} Z. He, Y. Wang, C. Zhi, Y. Hu, T.-P. Chen, L. Yin, Z. Chen, T. A. Wu, S. Ouyang, Z. Wang, J. Pei, J. McAuley, Y. Choi, and A. Pentland. MemoryArena: Benchmarking agent memory in interdependent multi-session agentic tasks. \emph{arXiv preprint arXiv:2602.16313}, 2026.

\bibitem{webarena} S. Zhou, F. F. Xu, H. Zhu, X. Zhou, R. Lo, A. Sridhar, X. Cheng, T. Ou, Y. Bisk, D. Fried, U. Alon, and G. Neubig. WebArena: A realistic web environment for building autonomous agents. In \emph{International Conference on Learning Representations}, 2024.

\bibitem{webshop} S. Yao, H. Chen, J. Yang, and K. Narasimhan. WebShop: Towards scalable real-world web interaction with grounded language agents. In \emph{Advances in Neural Information Processing Systems}, 2022.

\bibitem{swebench} C. E. Jimenez, J. Yang, A. Wettig, S. Yao, K. Pei, O. Press, and K. Narasimhan. SWE-bench: Can language models resolve real-world GitHub issues? In \emph{International Conference on Learning Representations}, 2024.

\bibitem{mem0} P. Chhikara, D. Khant, S. Aryan, T. Singh, and D. Yadav. Mem0: Building production-ready AI agents with scalable long-term memory. \emph{arXiv preprint arXiv:2504.19413}, 2025.

\bibitem{memgpt} C. Packer, S. Wooders, K. Lin, V. Fang, S. G. Patil, I. Stoica, and J. E. Gonzalez. MemGPT: Towards LLMs as operating systems. \emph{arXiv preprint arXiv:2310.08560}, 2023.

\bibitem{graphrag} D. Edge, H. Trinh, N. Cheng, J. Bradley, A. Chao, A. Mody, S. Truitt, and J. Larson. From local to global: A graph RAG approach to query-focused summarization. \emph{arXiv preprint arXiv:2404.16130}, 2024.

\bibitem{memorag} H. Qian, Z. Liu, P. Zhang, K. Mao, D. Lian, Z. Dou, and T. Huang. MemoRAG: Boosting long context processing with global memory-enhanced retrieval augmentation. In \emph{Proceedings of the ACM Web Conference}, pages 2366--2377, 2025.

\bibitem{reasoningbank} S. Ouyang, J. Yan, I.-H. Hsu, Y. Chen, K. Jiang, Z. Wang, R. Han, T. T. Le, S. Daruki, X. Tang, et al. ReasoningBank: Scaling agent self-evolving with reasoning memory. \emph{arXiv preprint arXiv:2509.25140}, 2025.

\bibitem{demem} M. Zou, Z. Guo, L. Liang, Z. Wang, Q. Wang, Q. Wen, I. King, L. Qu, and Z. Xu. Remember the decision, not the description: A rate-distortion framework for agent memory. \emph{arXiv preprint arXiv:2605.10870}, 2026.

\bibitem{msce} B. Tang, Y. Zhang, G. Zhuang, W. Wei, G. Zheng, L. Xie, Y. Tan, F. Xiong, Q. Yang, E. Chung, and Z. Li. From memory to skills: Evidence-grounded co-evolution governance for long-horizon LLM agents. \emph{arXiv preprint arXiv:2607.16621}, 2026.

\end{thebibliography}
\end{document}